\documentclass[10pt,twocolumn,letterpaper]{article}

\usepackage{titlesec}
\titlespacing*{\section}
{0pt}{5.5ex plus 1ex minus .2ex}{4.3ex plus .2ex}
\titlespacing*{\subsection}
{0pt}{5.5ex plus 1ex minus .2ex}{4.3ex plus .2ex}

\usepackage{wacv}
\usepackage{times}
\usepackage{epsfig}
\usepackage{amsmath}
\usepackage{amssymb}
\usepackage[ruled,vlined]{algorithm2e}
\usepackage{algpseudocode}
\usepackage[]{array}
\usepackage[]{booktabs}
\usepackage{caption}
\usepackage[]{multirow}
\usepackage{makecell}

\SetCommentSty{mycommfont}

\SetKwInput{KwInput}{Input}                % Set the Input
\SetKwInput{KwOutput}{Output}              % set the Output

\def\wacvPaperID{89} % Enter the WACV Paper ID here

\wacvfinalcopy % *** Uncomment this line for the final submission

\ifwacvfinal
\usepackage[breaklinks=true,bookmarks=false]{hyperref}
\else
\usepackage[pagebackref=true,breaklinks=true,colorlinks,bookmarks=false]{hyperref}
\fi

\begin{document}

%%%%%%%%% TITLE
\title{Defense-friendly Images in Adversarial Attacks: Dataset and Metrics for Perturbation Difficulty}

\author{Camilo Pestana\\
{\tt\small camilo.pestanacardeno@research.uwa.edu.au}
\and
Wei Liu\\
{\tt\small wei.liu@uwa.edu.au}
\and
David Glance\\
{\tt\small david.glance@uwa.edu.au}
\and
Ajmal Mian\\
{\tt\small ajmal.mian@uwa.edu.au}
\and
The University of Western Australia\\
35 Stirling Hwy, Crawley WA 6009, Australia\\
}

\maketitle
\thispagestyle{empty}
%Rev-4 More clear abstract. Reviewer 4 suggested a more clear abstract. It seems that the conclusion was more clear than the abstract%
% Rev-7. I think many of the corcerns of rev-7 could be answer here in the abstract and in the conclusion.
%%%%%%%%% ABSTRACT
\begin{abstract}
\textcolor{blue}{
%A problem that has plagued computer vision since its inception is that of biased datasets. An algorithm may show better results on the reported dataset than can be replicated on other datasets. Even when two algorithms are compared, their relative performance can vary depending on the dataset. 
% The above is too broad. Many of the problems have also been solved because of standard datasets and Kaggle challenges
}
Dataset bias is a problem in adversarial machine learning, especially in the evaluation of defenses. An adversarial attack or defense algorithm may show better results on the reported dataset than can be replicated on other datasets. Even when two algorithms are compared, their relative performance can vary depending on the dataset.
Deep learning offers state-of-the-art solutions for image recognition, but deep models are vulnerable even to small perturbations. Research in this area focuses primarily on adversarial attacks and defense algorithms. In this paper, we report for the first time, a class of robust images that are both resilient to attacks and that recover better than random images under adversarial attacks using simple defense techniques. Thus, a test dataset with a high proportion of robust images gives a misleading impression about the performance of an adversarial attack or defense. We propose three metrics to determine the proportion of robust images in a dataset and provide scoring to determine the dataset bias. We also provide an ImageNet-R dataset of 15000+ robust images to facilitate further research on this intriguing phenomenon of image strength under attack. Our dataset, combined with the proposed metrics, is valuable for unbiased benchmarking of adversarial attack and defense algorithms.

\end{abstract}

%%%%%%%%% BODY TEXT
\section{Introduction}

Convolutional Neural Networks (CNNs) are used in a wide range of computer vision tasks~\cite{krizhevsky2012ImageNet} such as object detection and image classification. However, in 2014, Szegedy et al.~\cite{szegedy2013intriguing} demonstrated that small perturbations in an input image can make a CNN model misclassify it. These perturbed images are called Adversarial Examples. As attacks are getting more sophisticated~\cite{Rony_2019_CVPR}, there is an increasing need for more robust models and better defenses~\cite{erichson2019jumprelu}. %Similarly, the importance of better metrics to measure robustness of models and validate defense techniques becomes more relevant.

Pestana et al.~\cite{pestana2020adversarial} suggested that depending on the subset of images selected from the ImageNet~\cite{krizhevsky2012ImageNet} validation set, results reported by different adversarial defenses (for ImageNet-based trained models) might differ remarkably. This is a significant problem that tarnishes the validity of current evaluations reported in Adversarial Defense research. To date, there are three main benchmark datasets that are used to evaluate the performance of adversarial attacks and defences in images, namely MNIST~\cite{lecun1998gradient}, CIFAR10~\cite{krizhevsky2009learning}, and ImageNet~\cite{krizhevsky2012ImageNet}. Of these datasets, ImageNet is the most complex given the number of classes (1,000) and the input image size (some models resize the images from ImageNet to 300$\times$300 pixels or more commonly 224$\times$224). MNIST and CIFAR10 datasets are commonly used to quickly demonstrate multiple attacks and defense techniques. However, these attacks do not transfer well to other datasets. 
 %and defenses do not always transfer well on MNIST and CIFAR10. 
 For example, the defense technique JumpReLU~\cite{erichson2019jumprelu} works well on CIFAR10 and MNIST but is not effective on ImageNet.
 
Hendrycks et al.~\cite{hendrycks2019natural} curated two datasets for ImageNet models that contain 7,500 hard-to-classify images or ``Naturally Adversarial Images" 
%This relates to Rev-4 Weaknesses Q-3%
and named this dataset as ImageNet-A (where `A' stands for adversarial). Their aim was to provide a new way to measure \textit{model} robustness and uncertainty.

On the other side of the spectrum, there exists a subset of images that is more difficult to perturb, i.e. "Naturally Robust Images". This was hinted to by Pestana et al.~\cite{pestana2020adversarial}. However, there is no readily available metrics to measure the \textit{data} robustness.

Our finding suggests that images that are more difficult to perturb are also easier to recover when using defenses, even if the defenses are weak. In this paper, we use the term  ``defense-friendly'' to describe the subset of images that are not only more robust but also recover more easily from an attack using defenses. To further demonstrate the existence of such a group of images, we curated a dataset of defense-friendly images.

\noindent The main contributions of our work are:
\\
\noindent(1) \textbf{Robust Dataset}: Existing datasets are not sufficiently challenging for evaluating adversarial attack/defense mechanisms. In this paper, a method for identifying robust images is proposed and used to construct such a dataset.

\noindent\textbf{For benchmarking:} We curated a dataset (ImageNet-R) containing easy to classify, robust to perturbation and defense-friendly images. This dataset will be released publicly to augment the ImageNet-A dataset. We use this dataset to validate the aforementioned claims about the shortcomings of existing datasets. 

\noindent\textbf{For Robustness Research:} Our dataset ImageNet-R  can inspire future research regarding the phenomenon of robust/defense-friendly images and create new lines of research.

\noindent\textbf{For Adversarial Research:} A dataset with a large proportion of robust images might give the wrong impression that the performance of a defense is better than it actually is. On the other hand, using a robust (hard-to-perturb) dataset might provide insights on how to improve existing attacks.

\noindent(2) \textbf{Robustness Metrics}: It is important to quantify the robustness of existing datasets, we therefore propose three metrics, namely ARD, AMP, and ADF score to measure the characteristics of a dataset with respect to adversarial attacks and defense algorithms. Given that those metrics are expensive to compute, a proxy method based on machine learning (i.e., predicting the robustness of an image) is introduced. We show the effectiveness of this proxy as a useful tool for creating unbiased benchmarks.

%\noindent(1) We show the existence of \textit{defense-friendly} images.

%\noindent(2) We conduct experiments to understand what makes those images more robust or defense-friendly.

%\noindent(3) We propose three metrics to evaluate datasets based on their robustness.

% Table generated by Excel2LaTeX from sheet 'Results-WACV'
\begin{table*}[t]
  \scriptsize
  \centering
  %\centering
  \setlength\tabcolsep{0.9pt}
  \caption{Comparison of Non-Robust (NR1) and Robust (R1) datasets accuracy (\%) under different attacks and  defenses. Images from R1 dataset are easier to defend not only for a single model but for all of them. Even though, the transferability of this property ``defense-friendly” is not 100\% between models, it shows that compared to a non-robust dataset (NR1), the likelihood of a robust image to recover from the wrong class prediction when using a defense is significantly higher. These robust images are also significantly more difficult to perturb and attacks would need a higher level of perturbation to be more effective. Higher perturbations levels compromise imperceptibility.}
    \begin{tabular}{clrrrrrrrrrrrrrrrrrrrrrrrrrrrrrr}
    \toprule
    \multirow{4}[8]{*}{Data} & \multicolumn{1}{c}{\multirow{4}[8]{*}{Models}} & \multicolumn{10}{c}{No Defense}                                     & \multicolumn{10}{c}{BaRT Defense}                                            & \multicolumn{10}{c}{ResUpNet Defense} \\
\cmidrule{3-32}          &       & \multicolumn{1}{c}{\multirow{3}[6]{*}{Clean}} & \multicolumn{3}{c}{FGSM} & \multicolumn{3}{c}{PGD} & \multicolumn{3}{c}{DDN} & \multicolumn{1}{c}{\multirow{3}[6]{*}{Clean}} & \multicolumn{3}{c}{FGSM} & \multicolumn{3}{c}{PGD} & \multicolumn{3}{c}{DDN} & \multicolumn{1}{c}{\multirow{3}[6]{*}{Clean}} & \multicolumn{3}{c}{FGSM} & \multicolumn{3}{c}{PGD} & \multicolumn{3}{c}{DDN} \\
\cmidrule{4-12}\cmidrule{14-22}\cmidrule{24-32}          &       &       & \multicolumn{3}{c}{$\epsilon$} & \multicolumn{3}{c}{$\epsilon$} & \multicolumn{3}{c}{n} &       & \multicolumn{3}{c}{$\epsilon$} & \multicolumn{3}{c}{$\epsilon$} & \multicolumn{3}{c}{n} &       & \multicolumn{3}{c}{$\epsilon$} & \multicolumn{3}{c}{$\epsilon$} & \multicolumn{3}{c}{n} \\
\cmidrule{4-12}\cmidrule{14-22}\cmidrule{24-32}          &       &       & \multicolumn{1}{c}{0.01} & \multicolumn{1}{c}{0.02} & 0.04  & \multicolumn{1}{|c}{0.01} & \multicolumn{1}{c}{0.02} & 0.04  & \multicolumn{1}{|c}{20}    & 40    & \multicolumn{1}{c}{60} &       & \multicolumn{1}{c}{0.01} & \multicolumn{1}{c}{0.02} & \multicolumn{1}{c}{0.04} & \multicolumn{1}{|c}{0.01} & \multicolumn{1}{c}{0.02} & \multicolumn{1}{c}{0.04} & \multicolumn{1}{|c}{20} & \multicolumn{1}{c}{40} & \multicolumn{1}{c}{60} &       & \multicolumn{1}{c}{0.01} & \multicolumn{1}{c}{0.02} & \multicolumn{1}{c}{0.04} & \multicolumn{1}{|c}{0.01} & \multicolumn{1}{c}{0.02} & \multicolumn{1}{c}{0.04} & \multicolumn{1}{|c}{20} & \multicolumn{1}{c}{40} & \multicolumn{1}{c}{60} \\
    \midrule
    \multirow{7}[14]{*}{\makecell{Non- \\ Robust \\ (NR1)}} & vgg   & 68.7  & 0     & 0     & 0     & 0     & 0     & 0     & 0     & 0     & 0     & 44.4  & 1.3   & 0.6   & 0     & 1.1   & 0.7   & 0.0   & 4.0   & 5.0   & 4.0   & 41.4  & 9.3   & 6.7   & 0.1   & 12.0  & 9.0   & 6     & 31.8  & 31.8  & 31.8 \\
\cmidrule{3-32}          & resnet & 81.2  & 3.1   & 6.2   & 3.1   & 0     & 0     & 0     & 0     & 0     & 0     & 62.7  & 4.5   & 8.7   & 4.9   & 5.9   & 4.1   & 2.0   & 8.0   & 8.0   & 11.0  & 61.0  & 31    & 28.1  & 28.1  & 28.0  & 14.0  & 6     & 47.3  & 47.3  & 47.3 \\
\cmidrule{3-32}          & densenet & 87.5  & 15.6  & 9.4   & 6.2   & 0     & 0     & 0     & 0     & 0     & 0     & 76.3  & 4.3   & 15.2  & 9.8   & 13    & 9.3   & 8.0   & 22.0  & 17.0  & 16.0  & 84.0  & 47    & 28.4  & 25.7  & 37.0  & 16.0  & 6     & 72.6  & 72.6  & 72.6 \\
\cmidrule{3-32}          & inception & 85.9  & 14.4  & 13.4  & 9.4   & 0     & 0     & 0     & 66.6  & 66.6  & 66.6  & 58.1  & 4.7   & 2.9   & 3.5   & 5.1   & 2.9   & 3.7   & 8.8   & 8.9   & 8.4   & 60.9  & 19.4  & 13.1  & 3.1   & 9.6   & 0     & 0     & 39.04 & 39.04 & 39.04 \\
\cmidrule{3-32}          & mobilenet & 85.9  & 12.8  & 9.1   & 7.2   & 0     & 0     & 0     & 73.4  & 73.4  & 73.4  & 34.1  & 4.1   & 2.9   & 1.2   & 4.4   & 1.9   & 1.7   & 5.3   & 6.9   & 7.0   & 40.6  & 22.5  & 1.1   & 0.0   & 15.6  & 0     & 0     & 34.4  & 34.4  & 34.4 \\
\cmidrule{3-32}          & shufflenet & 87.5  & 8.1   & 6.0   & 4.1   & 0     & 0     & 0     & 70.3  & 70.3  & 70.3  & 37.2  & 1.4   & 1.0   & 0.5   & 1.7   & 1.3   & 1.2   & 7.2   & 5.4   & 5.9   & 32.8  & 29.2  & 4.7   & 1.6   & 12.5  & 1.6   & 1.6   & 23.4  & 23.4  & 23.4 \\
\cmidrule{3-32}          & mnasnet & 90.6  & 11.2  & 3.4   & 4.1   & 0     & 0     & 0     & 78.1  & 78.1  & 78.1  & 39.1  & 4.1   & 3.2   & 1.7   & 2.9   & 2.9   & 1.7   & 7.3   & 7.1   & 6.8   & 40.6  & 28.8  & 9.4   & 1.6   & 20.3  & 7.8   & 3.1   & 34.4  & 34.4  & 34.4 \\
\midrule
\midrule
    \multirow{7}[14]{*}{\makecell{Robust \\ (R1)}} & vgg   & 89.8  & 83.3  & 82.2  & 79.5  & 54.3  & 26.3  & 11.2  & 83.2  & 83.2  & 83.2  & 89.8  & 76.9  & 73.7  & 72.8  & 55.4  & 33.8  & 19.9  & 81.6  & 80.1  & 80.3  & 61.1  & 71.5  & 71.1  & 70.4  & 69.2  & 65.8  & 59.9  & 62.8  & 62.8  & 62.8 \\
\cmidrule{3-32}          & resnet & 98.0  & 100   & 100   & 100   & 99.8  & 43.9  & 17.5  & 100   & 100   & 100   & 98.3  & 94.7  & 95.2  & 95.1  & 93.9  & 57.2  & 34.9  & 98.0  & 98.7  & 98.6  & 83.9  & 87.4  & 87.3  & 86.9  & 86.0  & 83.9  & 79.2  & 83.9  & 83.9  & 83.9 \\
\cmidrule{3-32}          & densenet & 99.2  & 100   & 100   & 100   & 100   & 55.5  & 22.1  & 100   & 100   & 100   & 99.2  & 97.0  & 97.1  & 96.9  & 95.9  & 67.0  & 40.9  & 99.3  & 98.7  & 99.0  & 90.7  & 92.8  & 92.8  & 92.0  & 90.6  & 88.4  & 83.5  & 90.7  & 90.7  & 90.7 \\
\cmidrule{3-32}          & inception & 100   & 97.9  & 97.9  & 97.1  & 87.2  & 64.5  & 42.9  & \multicolumn{1}{r}{98.1} & \multicolumn{1}{r}{98.1} & 98.1  & 98.3  & 94.0  & 93.0  & 94.0  & 86.8  & 72.7  & 52.9  & 96.7  & 97.2  & 96.5  & 86.9  & 90.5  & 90.4  & 90.3  & 89.0  & 87.1  & 83.3  & 86.5  & 86.5  & 86.5 \\
\cmidrule{3-32}          & mobilenet & 99.2  & 89.0  & 88.2  & 85.9  & 55.4  & 23.2  & 8.4   & 88.8  & 88.8  & 88.8  & 90.2  & 82.0  & 84.0  & 81.0  & 60.0  & 34.7  & 22.3  & 87.7  & 86.3  & 86.5  & 74.7  & 76.0  & 75.6  & 74.8  & 73.5  & 70.8  & 65.7  & 74.5  & 74.5  & 74.5 \\
\cmidrule{3-32}          & shufflenet & 99.1  & 82.4  & 80.8  & 77.6  & 41.3  & 16.6  & 5.8   & 82.5  & 82.5  & 82.5  & 87.4  & 76.0  & 75.0  & 71.0  & 48.4  & 28.7  & 17.7  & 80.8  & 78.6  & 79.3  & 65.0  & 70.2  & 69.8  & 69.0  & 66.7  & 62.7  & 56.7  & 65.8  & 65.8  & 65.8 \\
\cmidrule{3-32}          & mnasnet & 99.6  & 90.5  & 89.6  & 87.7  & 57.9  & 23.6  & 8.6   & 90.7  & 90.7  & 90.7  & 92.8  & 81.9  & 81.8  & 82.2  & 61.4  & 36.1  & 24.9  & 88.6  & 87.3  & 87.6  & 71.1  & 72.6  & 72.4  & 71.7  & 70.2  & 67.0  & 61.7  & 71.0  & 71.0  & 71.0 \\
    \bottomrule
    \end{tabular}%
  \label{tab:results_datasets}%
\end{table*}%

\section{Problem Definition}

\subsection{ Adversarial Attacks and Defenses}
Let  $x$ $\in$ $\mathbb{R}^{H \times W \times 3}$ denote the original image without perturbations, where H is height, W is width and there are three color channels, usually (R)ed, (G)reen and (B)lue. Let $y$ $\in$ $\mathbb{R}^n$ be the probabilities of the predicted labels. Given an image classifier $C:$ $x \rightarrow{\{1,2,...,n\}}$, e.g., for an ImageNet dataset \cite{imagenet_cvpr09} $n$ = 1,000, an untargeted attack aims to add a perturbation $p$ to $x$ to compute $x_{adv}$, such that $C(x)\neq C(x_{adv})$, where $p$ could be a grayscale image of the same size $p$ $\in$ $\mathbb{R}^{H \times W}$; or a colored image of the same size $p$ $\in$ $\mathbb{R}^{H \times W \times 3}$. The calculation $x_{adv} = x + p$ is constrained such that the perturbation in $x_{adv}$ is imperceptible for the human eye, e.g., $d(x, x_{adv}) \leq \epsilon$ for a distance function $d()$ and a small value $\epsilon$. In the context of adversarial attacks, the distance metric $d()$ is the $L_{p}$ norm of the difference between the original image $x$ and the adversarial image $x_{adv}$. In this paper, we consider attacks that optimise the $L_{2}$ and $L_{\infty}$ norms only. If the computed $x_{adv}$ satisfies the condition $C(x) \neq C(x_{adv})$ under the given set of constraints, the attack is considered successful. In addition, let $D()$ denote a defense function; if $C(x)\neq C(x_{adv})$ then $D()$ should ideally behave such that $C(D(x_{adv})) = C(x)$.

\subsection{Easy, Robust and Defense-friendly Images}
Given a classifier $C_{i}()$, an image $x$ is considered \textit{easy} if different classifiers trained on the same data, e.g., DenseNet121~\cite{huang2017densely}, Inception-v3~\cite{szegedy2016rethinking}, Vgg16~\cite{simonyan2014very}, agree with the prediction such that $C_{0}(x) =   C_{1}(x) = ... = C_{n}(x)$. An image $x$ is considered  \textit{$\epsilon$-robust} if its adversarial version $x_{adv}$ is classified correctly such that $C(x_{adv}) = C(x)$ for a small $\epsilon$ value. Therefore, an easy image can be considered a special case of the most general definition of robust images where $\epsilon $= 0. 
%I added this according to Wei's feedback
In our experiments, we found that $\epsilon$-robust images are defense friendly at higher $\epsilon$, where $\epsilon$ is the highest perturbation that an image can withstand without misclassification. Moreover, at the same perturbation level, $\epsilon$-robust images and epsilon defense friendly images are a complementary set, in that they do not have intersections. Therefore, images that are robust have a higher chance of recovering accuracy with correct class prediction at inference time. In other words, robust images are often ``defense-friendly". As shown in Table \ref{tab:results_datasets}, a test set with more robust images return better accuracy under the same defense mechanism.

\section{Related Work}
We briefly discuss popular adversarial attacks and defenses proposed in the literature that will be used in our experiments. We only focus on adversarial examples in the domain of image classification. In addition, techniques to obtain statistical features from images are discussed along with interpretability methods in Deep Learning.

\subsection{Attack Algorithms}
Gradient-based attacks are more powerful than Non-gradient-based attacks and usually less computationally expensive \cite{akhtar2018threat}, hence a defense against them is practically more meaningful. We use gradient-based attacks to test and compare our robust datasets and test the performance of some defenses using different sets of images.

\noindent \textbf{Fast Gradient Signed Method (FGSM)}:  This attack is one of the first computationally efficient single step adversarial attacks mentioned in the literature and introduced by Goodfellow et al.~\cite{goodfellow2014explaining}. FGSM is calculated like $x_{adv} = x + \varepsilon * \textrm{sign}(\nabla x J(\theta, x, y))$  where  $x_{adv}$ is the adversarial image and $x$ is the original image.  In the cost function $J(\theta, x, y)$, $\theta$ represents the network parameters and $y$ the ground truth label. Moreover, $\epsilon$ is used to scale the noise and is usually a small number (e.g. $\epsilon$=0.01). A sign function is applied to the gradients of the loss with respect to the input image to compute the final perturbation. 

\noindent \textbf{Projected Gradient Descent (PGD):} A stronger iterative version of FGSM is the PGD attack, which is considered one of the strongest attacks and used as a benchmark to measure the robustness of many defenses in the literature~\cite{madry2017towards}. It works similarly to FGSM, however, in this iterative version a small perturbation step $\alpha$ is applied in every step. The ``projection", for example, for an $L_{2}$ norm in a 2 dimensional space would mean to move a point to the closest point inside a circumference where the center is the origin (original image without perturbation). 

\noindent \textbf{Carlini and Wagner (C\&W):} Carlini and Wagner~\cite{carlini2017towards} introduced three attacks in response to one of the first adversarial defenses in the literature called Defensive Distillation. These attacks are hardly noticeable by humans, given that they constrain the $L_{p}$-norm. The C\&W attacks are shown to be highly effective against some types of defenses. A C\&W attack that is constrained by the $L_{2}$ norm is considered one of the strongest attacks in the literature \cite{rony2019decoupling}, however, this attack can be computationally expensive, often requiring thousands of iterations to calculate an adversarial image.

\noindent \textbf{Decoupling Direction and Norm (DDN):}
The DDN attack is an efficient method that was introduced by Rony et al. ~\cite{rony2019decoupling}. DDN optimizes the number of iterations needed while achieving state-of-the-art results, even comparable to C\&W $L_{2}$ constrained attacks.

\subsection{Defenses}
Recently, a number of studies have explored ideas to make Deep Neural Networks (DNN) more robust against adversarial attacks. Many of those defense methods are still unable to achieve true robustness to all adversarial inputs. Currently, the most effective defense strategies modify the DNN training process to improve robustness against adversarial examples ~\cite{shaham2018understanding}. However, they are trained to defend against specific attacks, limiting their real-world applications. In contrast, there is another research line on defenses which aims to be attack and model agnostic. This line preprocesses the images instead of modifying models or applying specific defenses to them. These two different approaches are:  defenses implemented in the training process ~\cite{stutz2019disentangling,taghanaki2019kernelized,liu2019rob} and defenses that are applied as an image denoising operation~\cite{raff2019barrage,xie2019feature}. Below we describe some of the defenses with respect to these two categories. However, our aim in this paper is to have a model and defense agnostic approach in the experiments. For that reason, our focus is mainly on the image preprocessing or image denoising techniques.

\subsubsection{\textbf{Adversarial Training}}
\vspace{-1mm}
 The aim of adversarial training is to increase the DNN's robustness by adding adversarial images to the training set~\cite{goodfellow2014explaining},~\cite{lyu2015unified},~\cite{shaham2018understanding}. These methods effectively enhance robustness against adversarial attacks, but they lack generalisation to unknown attacks ~\cite{tramer2017ensemble}. However, as demonstrated by ~\cite{narodytska2016simple} and \cite{papernot2017practical}, this type of defense is not effective to non-gradient-based attacks. 

\subsubsection{\textbf{Image Denoising}}
\vspace{-1mm}
The approach of pre-processing input examples to remove adversarial perturbations has the advantage of being model-agnostic. Moreover, it can be combined with any other defense strategies. Bhagoji et al. ~\cite{bhagoji2017dimensionality} proposed a Principal Component Analysis (PCA) method on images to reduce their dimentionality and therefore reduce noise. Alternatively, Das et el. ~\cite{das2017keeping} proposed leveraging JPEG compression as a pre-processing step for adversarial defenses. 
\\
The defense Barrage of Random Transformations (BaRT) combines different input transformation defenses in the literature such as Feature Squeezing ~\cite{xu2017feature}, Blurring Filters~\cite{cohen2019certified} and JPEG Compression~\cite{dziugaite2016study},~\cite{guo2017countering} into a pipeline. He et al.~\cite{he2017adversarial} showed that combining weak defenses does not create a stronger defense. However, Raff et al.~\cite{raff2019barrage} demonstrated that randomly selecting transformations from a big pool of transformations into a single barrage (BaRT) is a defense capable of resisting the strongest attacks such as PGD. This defense has a pool of 25 different transformations that can be randomly selected from the pool. The parameters for every transformation are selected randomly as well as the order $k$ of the number of transformations selected. Then, once they are combined in a single pipeline the transformations are applied sequentially. By evaluating this defense, we are indirectly testing and comparing the performance against 25 different input transformations.
\\
\noindent ResUpNet defense~\cite{pestana2020adversarial} exploits the insight that the Y channel from the YCbCr is more relevant in terms of adversarial perturbation and focuses on recovering the Y-channel in that specific color space before converting it back to RGB. This defense is reported to achieve better than the BaRT defense.

\subsection{Image Features}
\vspace{-1mm}

Convolutional Neural Networks (CNNs) have become a very popular approach to extract features from images. However, for image analysis other statistical techniques also exist such as the grey level co-occurence matrix \cite{haralick1973textural} to measure the texture of an image. Ideally, in texture analysis the aspects of an image should be rotationaly invariant. A common practice is to calculate the grey co-occurence matrix using different angles (e.g. 0, 45, 90, 135)~\cite{8053537}. Hence, for our experiments we calculate Grey Level Co-occurence Matrices (GLCM) rotating the image at the 4 different angles mentioned. From those GLCM it is possible to extract the texture properties below:

\vspace{1mm}
\noindent1) \textbf{Contrast} refers to the calculation of the intensity contrast between a pixel a its neighbors i.e.~$\sum_{i,j=0}^{N-1} P_{i,j}(i-j)^{2}$.

\vspace{1mm}
\noindent2) \textbf{Dissimilarity} is the variation of grey-level pairs computed as $\sum_{i,j=0}^{N-1}P_{i,j}|i-j|$.

\vspace{1mm}
\noindent3) \textbf{Homogeneity} measures the tightness of distribution of the elements in the GLCM to the GLCM diagonal, $\sum_{i,j=0}^{N-1} \dfrac{P_{i,j}}{1+(i-j)^{2}}$.

\vspace{1mm}
\noindent4) \textbf{Angular Second Moment (ASM)} is a measure of the homogeneity of an image computed as $\sum_{i,j=0}^{N-1}\ P^{2}_{i,j}$.

\vspace{1mm}
\noindent5) \textbf{Energy} refers to the information related to image homogeneity. It has a low value if the pairs are similar and high values otherwise and is computed as $\sqrt{\sum_{i,j=0}^{N-1}\ P^{2}_{i,j}}$.

\vspace{1mm}
\noindent6) \textbf{Correlation} measures linear dependency between pixels. It has high values when the pixels are uniformly distributed and is computed as $\sum_{i,j=0}^{N-1} P_{i,j}
    \dfrac{(i- \mu_{i}) (j-\mu_{j})}{\sqrt{(\sigma_{i}^{2})(\sigma_{j}^{2})}}.
$

\section{Defense-friendly Dataset}

While evaluating the performance of adversarial attacks and defenses for ImageNet, it is a common practice to take a subset of images from ImageNet to test the performance of different techniques (attacks or defenses). However, some subsets of images perform much better or worse than a randomly selected subset~\cite{pestana2020adversarial}. This can lead to incorrect assumptions or conclusions with regards to the performance of models or defenses implemented. In this section, we explain the process we used to collect a set of images, outside of the ImageNet validation dataset, that are more resilient to attacks than randomly selected images and recover easily when using a defense (defense-friendly images). The source code and datasets will be made publicly available at https://github.com/elcronos/Defense-Friendly

\subsection{Implementation Details}
We used the ResUpNet~\cite{pestana2020adversarial} and BaRT~\cite{raff2019barrage} in the Pytorch~\cite{paszke2017automatic} library. In addition, AdverTorch~\cite{ding2019advertorch} was used to create adversarial images for three different classifiers Vgg16~\cite{simonyan2014very}, ResNet50~\cite{he2016deep}, and DenseNet121~\cite{huang2017densely}. Those models have unique architecture mechanisms and differ in the number of layers, which makes them a good benchmark to test the defenses using different datasets. For $L_\infty$ constrained attacks such as FGSM and PGD, we use $\epsilon$ values of 0.01, 0.02 and 0.04 for images with values in the range [0,1]. In the case of DDN attack, instead of $\epsilon$, we use the number of iterations $n$ as the hyper-parameter, with values 20, 40, and 60.

\subsection{Curating a Defense-friendly dataset}

To demonstrate the existence of defense-friendly images and to compare their performance with randomly selected images, we curated a dataset of defense-friendly images.
For this experiment, we used pre-trained ImageNet classifiers. We decided to create a new dataset using Flickr to prevent any data leakage  between the training and test datasets. In machine learning data leakage occurs because of the use of information in the model training process which is not available at prediction time. As a consequence, the predictive scores are overestimated.  We assume that very few or none of the images we collected are included in either the training ImageNet dataset or test dataset. Since ImageNet was collected scrapping the web in 2012, and we only collected images from the last two years from the Flickr API, we assume that the probability of having repeated images remains extremely low.

Flickr is a website that has millions of images shared by users. We used the Flickr API to download 1,000 images per each ImageNet class. The API uses a keyword system to download the images. The keywords used to download the images will also be shared in our GitHub repository.

\begin{figure*}[t!]
   \centering
   \includegraphics[width=\textwidth]{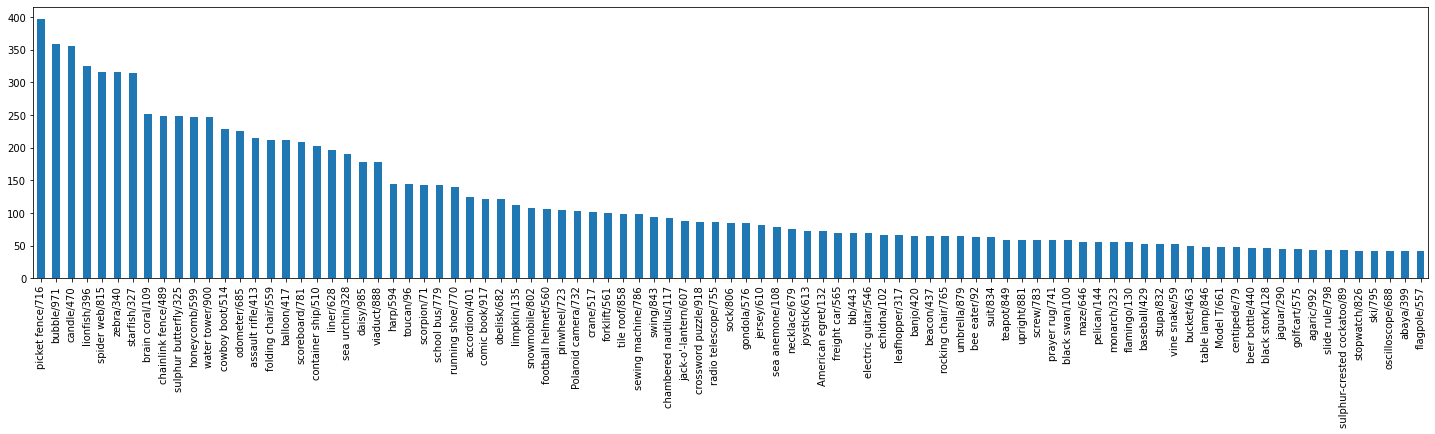} 
   \caption{
   This figure shows the per class frequency from the newly collected dataset (15,554 images) with the label and index number according to the ImageNet dataset. The figure shows only the classes that have at least 30 occurrences arranged in descending order.
   }
\label{fig:top_labels}
\end{figure*}

We used DenseNet-121, ResNet-50 and Inception-v3 pretrained models to further filter these images. We only used the image where all the three models correctly predict and agree on the label. From a million images that we originally collected using the Flickr API, only 434,337 (43\%) remained after this filtering process. The fact that different models are able to correctly classify the same image means that the image is robust enough when there is no perturbation (Easy Images). From the three different attacks used, we decided to use PGD attack for the rest of the experiments given that it is the strongest attack as shown in Table \ref{tab:accuracynodefense}.
%Rev-4 Q-7%
Moreover, we use ResUpNet and BaRT  defenses to find the defense-friendly images. %Ajmal: ResUpNet & Bart references?

For the easy images subset, a PGD attack was implemented per image using $\epsilon$ values of 0.01, 0.02 and 0.04 for each model. We defined those adversarial images as \textit{$\epsilon$-robust} if they were still classified correctly by the three models without using any defense algorithm. Depending on the amount of perturbation, and the ability of the image to retain their accuracy despite the added perturbation, some images are more robust than others. From the 434,337 robust images collected, only 15,554 images were able to retain the original prediction across different classifiers for adversarial attacks with an $\epsilon=0.01$. Those images were used to create the new robust dataset. 

\begin{figure*}[htbp]
   \centering
   \includegraphics[width=\linewidth,height=1.4in]{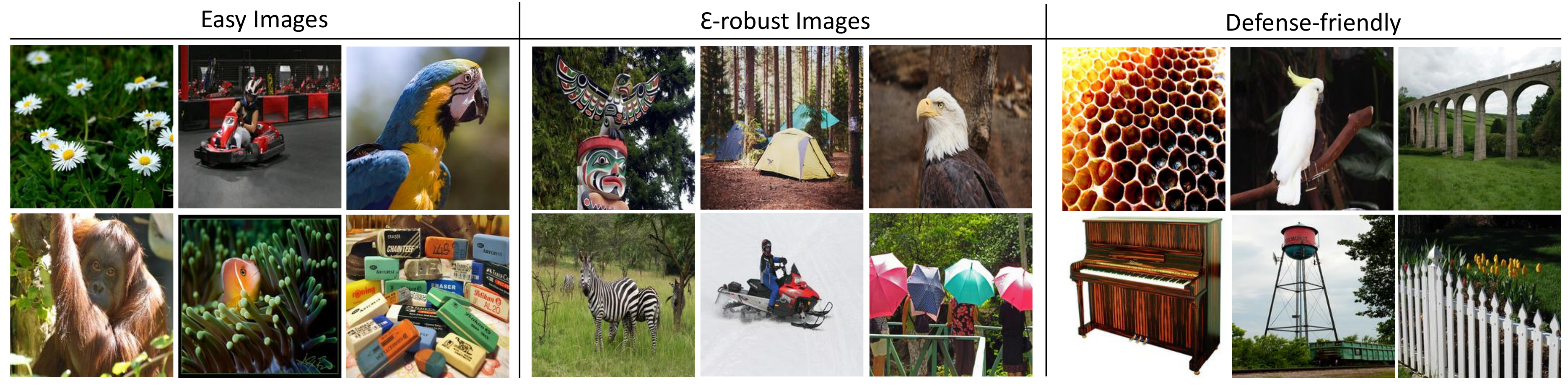}
   \caption{Sample of images from our curated dataset ImageNet-R for the categories: Easy, $\epsilon$-robust and Defense-friendly images.
   }
\label{fig:images}
\vspace{-5mm} 
\end{figure*}

%Ajmal: In Fig.3 call C as easy images otherwise C and D sound similar.

With extensive experimentation, we found that those images that are more \textit{robust} (can stand greater levels of perturbation in an attack) are also in general more \textit{defense-friendly}. Therefore, we consider defense-friendly images as a special case of robust images which makes them easier to recover from an attack. To better understand the difficulty of a dataset in regards to the attacks and defenses used, we introduced three dataset metrics that are further explained in Section 5.3. Figure \ref{fig:top_labels} shows the frequency of classes (according to the ImageNet labels), which has the greater number of correct samples in descending order with at least 30 occurrences. For this robust subset, the mean frequency of a the classes is 25.75 and the median 7. There are 604 classes with at least one image and the remaining 396 classes contain no image at all. A random sample of images corresponding to the three categories from our ImageNet-R dataset can be seen in Figure \ref{fig:images}.

\section{Performance Evaluating}

%Rev-4 Q-6 moved table by reviewer suggestion%
\begin{table}[h]
  %\centering
  \setlength\tabcolsep{1.15pt} % reduce blank spaces to half 3 pt (default 6 pts to fit table)
  \scriptsize
  \vspace*{-3mm}
  \caption{Accuracy of datasets with no defense:
  Datasets (D 1-3) get very similar results with very low standard deviation between them. However, datasets `Robust' (R 1-3) consistently get better results than the randomly selected datasets. Values given in percentage and all datasets have a total of 5,000 images.}
    \begin{tabular}{|c|l|r|r|r|r|r|r|r|r|r|r|}
    \hline
    {} &
    \multicolumn{1}{c|}{Model} & \multicolumn{1}{c|}{} & \multicolumn{3}{c|}{FGSM} & \multicolumn{3}{c|}{PGD} & \multicolumn{3}{c|}{DDN} \\
    &       &    \multicolumn{1}{c|}{$\epsilon$=0}   & \multicolumn{1}{c|}{$\epsilon$=0.01} & \multicolumn{1}{c|}{$\epsilon$=0.02} & \multicolumn{1}{c|}{$\epsilon$=0.04} & \multicolumn{1}{c|}{$\epsilon$=0.01} & \multicolumn{1}{c|}{$\epsilon$=0.02} & \multicolumn{1}{c|}{$\epsilon$=0.04} & \multicolumn{1}{c|}{$n$=20} & \multicolumn{1}{c|}{$n$=40} & \multicolumn{1}{c|}{$n$=60} \\
    \hline 
    {D1} & vgg   & 71.8 & 4.1 & 4.2 & 4.6 & 0.0 & 0.0 & 0.2 & 1.4 & 0.9 & 0.8 \\
         & resnet & 75.9 & 7.4 & 7.6 & 9.6 & 0.0 & 0.0 & 0.0 & 0.4 & 0.2 & 0.1 \\
         & dense & 74.6 & 4.2 & 4.3 & 5.6 & 0.4 & 0.2 & 0.0 & 0.3 & 0.1 & 0.1 \\
    \hline
    {D2} & vgg   & 70.0 & 4.0 & 3.7 & 4.5 & 0.5 & 0.3 & 0.2 & 1.3 & 0.1 & 0.1 \\
         & resnet & 74.8 & 6.7 & 6.3 & 8.0 & 0.0 & 0.0 & 0.0 & 0.3 & 0.1 & 0.1 \\
         & dense & 73.0 & 3.4 & 3.4 & 4.9 & 0.1 & 0.0 & 0.0 & 0.2 & 0.8 & 0.1 \\
    \hline
    {D3} & vgg   & 71.8 & 4.7 & 4.3 & 4.8 & 0.7 & 0.5 & 0.2 & 1.4 & 1.1 & 1.0 \\
         & resnet & 76.3 & 7.8 & 7.6 & 9.4 & 0.1 & 0.0 & 0.0 & 0.6 & 0.3 & 0.2 \\
         & dense & 74.6 & 4.0 & 3.7 & 5.4 & 0.1 & 0.0 & 0.0 & 0.3 & 0.2 & 0.1 \\
    \hline
    {R1} & vgg   & 99.2 & 83.6 & 82.5 & 79.2 & 54.1 & 26.1 & 10.9 & 83.2 & 83.2 & 83.2 \\
         & resnet & 100 & 100 & 100 & 100 & 99.8 & 45.0 & 17.7 & 100 & 100 & 100 \\
         & dense & 100 & 100 & 100 & 100 & 100 & 54.6 & 22.9 & 100 & 100 & 100 \\
    \hline
    {R2} & vgg   & 99.3 & 83.6 & 82.3 & 79.8 & 54.1 & 26.2 & 11.2 & 83.6 & 83.6 & 83.6 \\
         & resnet & 100 & 100 & 100 & 100 & 99.8 & 44.0 & 17.8 & 100 & 100 & 100 \\
         & dense & 100 & 100 & 100 & 100 & 100 & 54.9 & 23.2 & 100 & 100 & 100 \\
    \hline
    {R3} & vgg   & 99.3 & 83.6 & 82.3 & 79.8 & 54.1 & 26.2 & 11.2 & 83.6 & 83.6 & 83.6 \\
         & resnet & 100 & 100 & 100 & 100 & 99.8 & 44.0 & 17.8 & 100 & 100 & 100 \\
         & dense & 100 & 100 & 100 & 100 & 100 & 54.9 & 23.2 & 100 & 100 & 100 \\
    \hline
    \end{tabular}%
  \label{tab:accuracynodefense}%
\end{table}%

\subsection{Evaluation for Adversarial Attacks}
The idea in this section is to compare how the accuracy of different models is affected when using different attacks on a ``normal" dataset (randomized subset from ImageNet) and the new dataset ImageNet-R. In this section, we present the results from our experiments which further support our claim that some images are more difficult to perturb or more robust. We observe a significant difference in the accuracy between the new dataset and a ``normal" dataset under different attacks for ImageNet-based models. We randomly split robust images into three groups R (1-3). The same process was done for the normal datasets D (1-3). Table \ref{tab:accuracynodefense} shows that robust datasets have more images with 100\% or nearly 100\% accuracy on non-perturbed images, and they are also much more resilient to attacks for small perturbations $\epsilon$ in comparison to randomly selected datasets.

\subsection{Evaluation for Adversarial Defenses}
\vspace{-1mm}
Similar to the evaluation of adversarial attacks, a random dataset and the easy images (including Robust Images and Defense-friendly) dataset will be compared. If our hypothesis and claim hold true, those images that are more robust or difficult to misclassify should also be more defense-friendly or easier to recover when using a defense.

As can be seen in Table \ref{tab:results_datasets}, defense-friendly images are more resilient to attacks and they also recover better when using a defense. It also shows that for non-defense friendly images, the percentage of accuracy recovery while using a defense is much lower than a more robust dataset.

\vspace{-1mm}
\section{Metrics for Quantifying Dataset Robustness}%Robust vs Non-Robust Images Classification}

%\subsection{}
\vspace{-1mm}

\begin{figure}[t]
   \centering
   \includegraphics[width=\linewidth,height=1.8in]{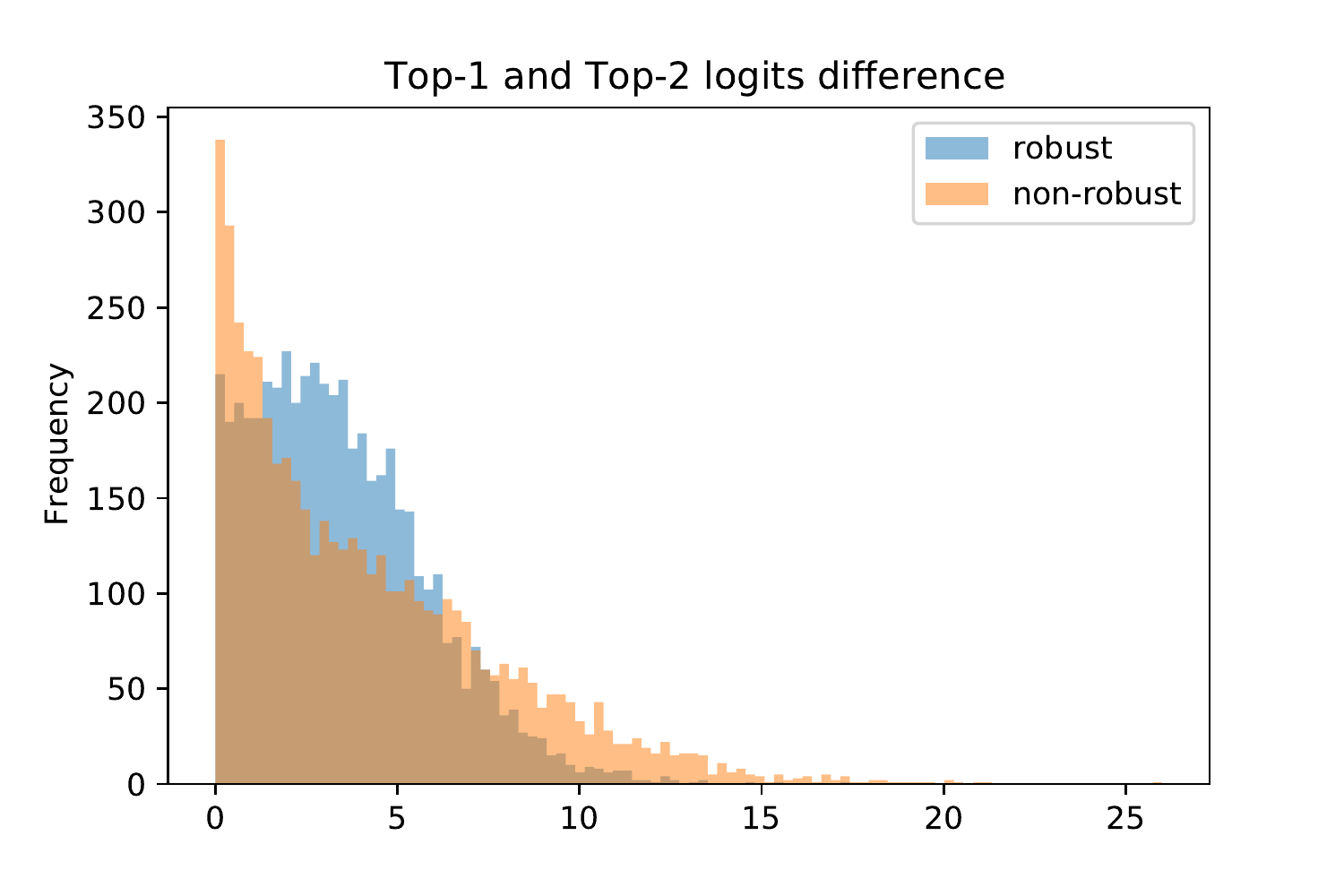}
   \caption{Distribution of the $\ell_2$ distance between the logit scores for the top-1 and top-2 predictions for a pre-trained ImageNet model (ResNet50) evaluated on a robust dataset and a randomized non-robust dataset. The third color in the figure shows their intersection. The difference between top-1 and top-2 logits is not always greater in robust images. In many cases it can also happen that the $\ell_2$ distance in the logits is similar in robust and non-robust.}
   \vspace{-3mm}
\label{fig:distribution}
\end{figure}

To measure the overall robustness of a dataset, we propose three different metrics, namely, ARD, AMP and ADF.
A general convention has been to use the difference between the top 2 predictions or logit scores as a confidence metric which is not reliable in general. We validate this by comparing the $\ell_2$ difference between the top 2 logit scores on ResNet50 for a pair of robust and non-robust datasets. Figure \ref{fig:distribution} shows our results. We can see that many non-robust images have a greater distance than that of robust images. On the other hand, our method uses a combination of three metrics that are more reliable and are designed to measure possible bias globally in a dataset as a whole rather than locally at individual images.% We expect that our ImageNet-R dataset will ignite future research on how to find better and more efficient metrics of robustness. However, the overall logic of the 3 proposed metrics and what every metric implies does not change even if a different heuristic is used to determine the score.}

% Rev-3 weaknesses Q-2
% New experiment. We discussed this before.

\noindent {\bf Adversarial Robust Dataset (ARD) Score:} This score provides the proportion of images in a dataset (range [0, 1]) that are robust given an attack $atk$  with a perturbation $\epsilon$ and a model $M$.

\noindent {\bf Adversarial Minimum Perturbation (AMP) score:} This metric computes the minimum $\epsilon$ perturbation in which a proportion of the images $P$ in a dataset $D$ are not able to resist an specific attack $atk$ for the model $M$. The higher the $\epsilon$ value returned by the function $AMP$, the easier the image dataset is to defend.

\noindent {\bf Adversarial Defense-friendly (ADF) Score:} This score finds the proportion of images in a dataset that are recovered using a defense for a specific attack with a small $\epsilon$.

\begin{algorithm}[h]
  \scriptsize
  \DontPrintSemicolon
  \caption{ARD Score}
  \KwInput{An adversarial attack $atk$, perturbation $\epsilon$, dataset $D$, model $M$} 
  \KwOutput{ARD Score denoted by $score$}
  \SetKwProg{CalculateARDScore}{CalculateARDScore}{}{}
  \CalculateARDScore{($atk,\epsilon, D, M$)}{
  $R \gets [ ]$ \tcp{Empty list for Robust Images}
  \tcp{For each image $i$ in dataset $D$}
  \ForEach{$i \in D$}{
    \tcp{Generate Adversarial Image}
    $i_{adv} \gets atk(i,\epsilon,M)$ \\
    \If{$M(i) = M(i_{adv})$}{
        $R$.insert($i$)
    }
  }
  $score$ = $\dfrac{Length(R)}{Length(D)}$ \\
  \KwRet{$score$}
  }
\end{algorithm}

\begin{algorithm}[h]
\DontPrintSemicolon
  \scriptsize
  \caption{AMP Score}
  \KwInput{An adversarial attack $atk$, perturbation $\alpha$, dataset $D$, model $M$, threshold $P$} 
  \KwOutput{AMP Score denoted by $\epsilon$}
  \SetKwProg{CalculateAMPScore}{CalculateAMPScore}{}{}
  \CalculateAMPScore{($atk,\epsilon, D, M, P$)}{
  \tcp{Empty list of Non-Robust Images}
  $ !R \gets [ ]$ \\
  $found_{\epsilon}  \gets False$ \\
  $\epsilon \gets 0$ \tcp{Initialize $\epsilon$ as zero}
  \While{$found_{\epsilon}$ = $False$}{
    \tcp{For each image $i$ in dataset $D$}
    \ForEach{$i \in D$}{ 
        \tcp{Generate Adversarial Image}
        $i_{adv} \gets atk(i,\epsilon,M)$ \\
        \If {$M(i) \neq M(i_{adv})$}{
            $!R$.insert($i$)
        }
    }
    $A \gets \dfrac{Length(!R)}{Length(D)}$
    
    \If{ $A \geq P$ }{
    $found_{\epsilon}  \gets True$
    }
    \Else{
    $\epsilon \gets \epsilon + \alpha$
    }
  }
  \KwRet{$\epsilon$}
  }
\end{algorithm}

\begin{algorithm}[h]
  \scriptsize
  \DontPrintSemicolon
  \caption{ADF Score}
  \KwInput{An adversarial attack $atk$, perturbation $\epsilon$, dataset $D$, model $M$} 
  \KwOutput{ADF Score denoted by $score$}
  \SetKwProg{CalculateADFScore}{CalculateADFScore}{}{}
  \CalculateADFScore{($atk,\epsilon, D, M$)}{
  \tcp{Empty list of Defense-friendly Images}
  $DF \gets [ ]$ \\
  \tcp{For each image $i$ in dataset $D$}
  \ForEach{$i \in D$}{
    \tcp{Generate Adversarial Image}
    $i_{adv} \gets atk(i,\epsilon,M)$ \\
    \If{$M(i) = M(i_{adv})$}{
        $DF$.insert($i$)
    }
  }
  $score$ = $\dfrac{Length(DF)}{Length(D)}$ \\
  \KwRet{$score$}
  }
\end{algorithm}

\noindent We created  6 different subsets of images (robust and non-robust) with images randomly selected from the collected dataset (Section 4). The results from the ARD, AMP and ADR Score applied to those datasets are shown in Table~\ref{tab:metricsresults}. 
%This relates with Rev-3 Weaknesses Q-1%
 Despite some models performing better, in general, robust images have a higher ADF score. We can also observe in the results that there are some transferability of robustness when we test on different models.
 
% Table generated by Excel2LaTeX from sheet 'ARD_Score'
\begin{table}[htbp]
  \setlength\tabcolsep{1.15pt} % reduce blank spaces to half 3 pt (default 6 pts to fit table)
  \scriptsize
  %This relates with Rev-3 Weaknesses Q-1%
  \caption{ARD, AMP and ADF scores for 6 different datasets, each containing 5,000 randomly selected images from our curated dataset. We used PGD attack for this experiment. Robust datasets (R1-R3) obtain better scores for each metric while Non-robust (NR1-NR3) datasets have significantly lower scores.}
  \vspace{1mm}
    \begin{tabular}{|c|l|r|r|r|r|r|r|r|r|r|r|r|r|r|}
    \hline
    Data & \multicolumn{1}{c|}{{Models}} & \multicolumn{3}{c|}{{ARD Score}} 
    & \multicolumn{4}{c|}{{AMP Score}} & \multicolumn{6}{c|}{ADF Score} \\
         &      & \multicolumn{3}{c|}{} & \multicolumn{4}{c|}{}& \multicolumn{3}{c|}{BaRT} & \multicolumn{3}{c|}{ResUpNet} \\
         &      & \multicolumn{3}{c|}{$\epsilon$} & \multicolumn{4}{c|}{Threshold} & \multicolumn{3}{c|}{$\epsilon$} & \multicolumn{3}{c|}{$\epsilon$} \\
    \hline
          &       & 0.01  & 0.02  & 0.04  & 50  & 70  & 80  & 90  & 0.01  & 0.02  & 0.04  & 0.01  & 0.02  & 0.04 \\
    \hline
    {NR1} & vgg   & 0.03  & 0.01  & 0.01  & 0.00  & 0.01  & 0.01  & 0.01  & 0.06  & 0.04  & 0.02  & 0.16  & 0.14  & 0.10 \\
         & resnet & 0.04  & 0.01  & 0.00  & 0.00  & 0.01  & 0.01  & 0.01  & 0.11  & 0.07  & 0.04  & 0.35  & 0.31  & 0.27 \\
         & dense & 0.07  & 0.01  & 0.00  & 0.01  & 0.01  & 0.01  & 0.01  & 0.15  & 0.08  & 0.04  & 0.38  & 0.31  & 0.25 \\
    \hline
    {NR2} & vgg   & 0.03  & 0.01  & 0.01  & 0.00  & 0.01  & 0.01  & 0.01  & 0.08  & 0.05  & 0.02  & 0.27  & 0.23  & 0.19 \\
         & resnet & 0.03  & 0.00  & 0.00  & 0.01  & 0.01  & 0.01  & 0.01  & 0.12  & 0.07  & 0.04  & 0.33  & 0.29  & 0.24 \\
         & dense & 0.08  & 0.02  & 0.00  & 0.01  & 0.01  & 0.01  & 0.01  & 0.18  & 0.09  & 0.07  & 0.41  & 0.34  & 0.27 \\
    \hline
    {NR3} & vgg   & 0.04  & 0.01  & 0.01  & 0.00  & 0.01  & 0.01  & 0.01  & 0.07  & 0.04  & 0.03  & 0.27  & 0.24  & 0.20 \\
         & resnet & 0.03  & 0.00  & 0.00  & 0.00  & 0.01  & 0.01  & 0.01  & 0.11  & 0.06  & 0.03  & 0.22  & 0.18  & 0.14 \\
         & dense & 0.07  & 0.01  & 0.00  & 0.01  & 0.01  & 0.01  & 0.01  & 0.17  & 0.10  & 0.06  & 0.42  & 0.35  & 0.27 \\
    \hline
    {R1} & vgg   & 0.51  & 0.25  & 0.11  & 0.01  & 0.01  & 0.01  & 0.01  & 0.56  & 0.36  & 0.22  & 0.66  & 0.63  & 0.57 \\
         & resnet & 1.00  & 0.45  & 0.18  & 0.02  & 0.02  & 0.02  & 0.03  & 0.98  & 0.62  & 0.37  & 0.85  & 0.82  & 0.77 \\
         & dense & 1.00  & 0.55  & 0.23  & 0.02  & 0.03  & 0.03  & 0.03  & 0.98  & 0.70  & 0.45  & 0.89  & 0.87  & 0.84 \\
    \hline
    {R2} & vgg   & 0.53  & 0.25  & 0.11  & 0.01  & 0.01  & 0.01  & 0.01  & 0.59  & 0.36  & 0.21  & 0.68  & 0.66  & 0.59 \\
         & resnet & 1.00  & 0.45  & 0.19  & 0.02  & 0.02  & 0.02  & 0.03  & 0.98  & 0.63  & 0.39  & 0.85  & 0.84  & 0.78 \\
         & dense & 1.00  & 0.55  & 0.22  & 0.02  & 0.03  & 0.03  & 0.03  & 0.99  & 0.71  & 0.42  & 0.91  & 0.89  & 0.83 \\
    \hline
    {R3} & vgg   & 0.53  & 0.26  & 0.11  & 0.01  & 0.01  & 0.01  & 0.01  & 0.57  & 0.36  & 0.21  & 0.68  & 0.65  & 0.59 \\
         & resnet & 1.00  & 0.44  & 0.18  & 0.02  & 0.02  & 0.02  & 0.02  & 0.97  & 0.59  & 0.37  & 0.87  & 0.83  & 0.79 \\
         & dense & 1.00  & 0.57  & 0.23  & 0.02  & 0.03  & 0.03  & 0.03  & 0.99  & 0.70  & 0.43  & 0.91  & 0.90  & 0.84 \\
    \hline
    \end{tabular}%
  \label{tab:metricsresults}%
  \vspace{5mm}
\end{table}

\begin{figure}[t]
   \centering
   \includegraphics[width=\linewidth, height=1in]{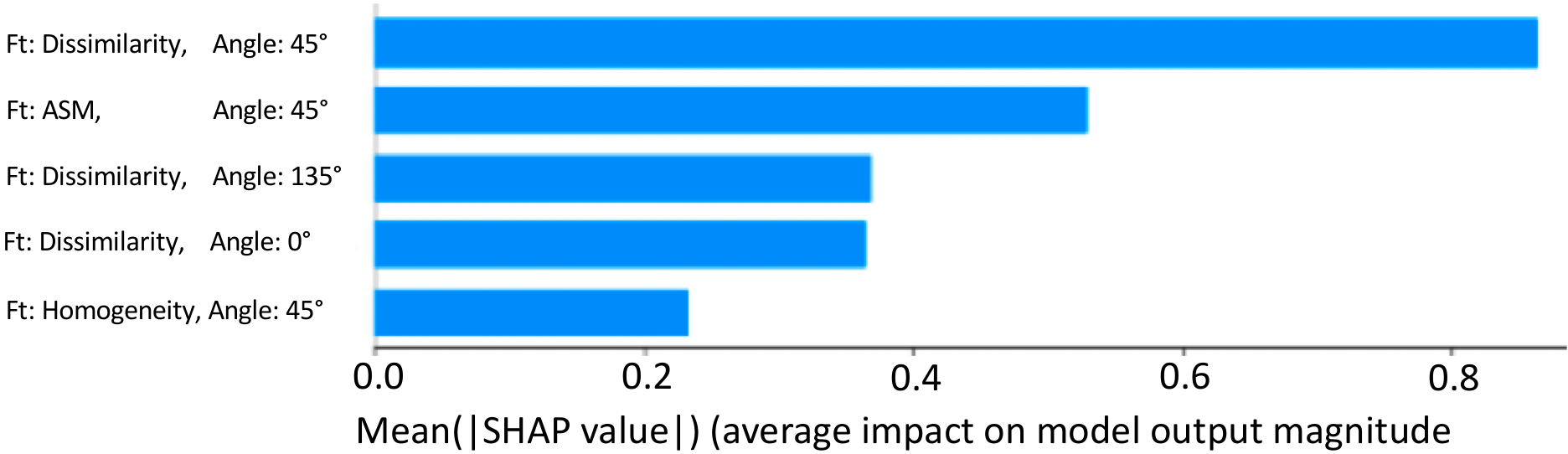}
   \caption{
   This figure shows a summary SHAP plot displaying the top 5 most important features (extracted using GLCM for the best performing ML model in Table \ref{tab:models}) and their contribution for every sample. This plot sorts features by the sum of SHAP value magnitudes over all samples. In addition, it shows the distribution of the impacts of every features with respect to the output. As a first insight, the model finds more useful features from distance 1 given in pixels. Moreover, the most significant texture feature seems to be dissimilarity (3 out of 5 features in the top 5 features) followed by the ASM feature using angle an angle of 45$^{\circ}$. A higher value of dissimilarity indicates that an image is more likely to be a non-robust image. On the contrary, the lower the dissimilarity in the image the higher the likelihood of being a robust image.
   }
\label{fig:shapplot}
\end{figure}

Initially, to differentiate robust images from non-robust ones, we use three losses (corresponding to the three models DenseNet, ResNet and Inception) and an epsilon value as a parameter to calculate the adversarial perturbations for that image. If the adversarial image resists a small $\epsilon$ value, we consider the image to be $\epsilon$-robust. However, this method is computationally expensive and hence, there is a need for a metric that can be computed more efficiently. One plausible solution is to use the first metric as the `ground truth’ labels and train a second classifier to classify the image as `robust’ or `non-robust’. The robust class is relative to a minimum level of perturbation $\epsilon$ which must be provided to create the `ground truth' labels.
%This relates with Rev-3 Weaknesses Q-1%

We used the GLCM features (see Section 3.3)  to extract statistical features from the 15,554 images that are considered robust and then randomly selected the same number of images from the non-robust pool to get a balanced dataset. A binary classifier (robust vs non-robust) was trained with the statistical features extracted using GLCM \cite{8053537}. Figure \ref{fig:shapplot} illustrates the importance of the statistical features such as dissimilarity in that binary classifier.
ImageNet-trained CNNs are biased towards texture \cite{geirhos_imagenet-trained_2019}, therefore, we want to analyse texture and determine if it is possible to extract texture elements that could indicate when an image is more robust than others. Next, a CNN model is trained in the classification problem (robust vs non-robust) to compare the results of the statistical approach versus the Deep Learning approach.

% Table generated by Excel2LaTeX from sheet 'models-results'
\begin{table}[t]
  \centering
  \setlength\tabcolsep{1pt} % reduce blank spaces to half 3 pt (default 6 pts to fit table)
  \scriptsize
  \vspace*{-3mm}
  \caption{Robust vs Non-robust classification of images. Before classifying the actual contents of an image, these classifiers determine whether an image is robust or not i.e. whether it is likely to be correctly classified by a deep model or not. GLCM features were extracted from the Y-channel of the YCbCr color space~\cite{pestana2020adversarial}.}
  \vspace{1mm}
    \begin{tabular}{|l|l|r|r|r|}
    \hline
     Model & Type Feature & \multicolumn{1}{l|}{Accuracy} & \multicolumn{1}{l|}{Precision} & \multicolumn{1}{l|}{~~Recall~} \\
    \hline
    Logistic Regression & Statistical GLCM & 66.1 & 69.6 & 54.9 \\
    Random Forest & Statistical GLCM & 75.0 & 74.6 & 74.5 \\
    SVM   & Statistical GLCM & 74.0 & 73.9 & 72.9 \\
    ResNet34 (Finetuned) & Deep Learning / CNN &    87.9   &   83.0    & 92.1 \\
    ResNet152 (Finetuned) & Deep Learning / CNN &    88.6   &   85.3    & 91.3 \\
    DenseNet121 (Finetuned) & Deep Learning / CNN &   88.0    &   84.1    & 92.2 \\
    \hline
    \end{tabular}%
  \label{tab:models}%
\end{table}%

For the statistical features, we train a simple Logistic Regression (baseline), a Random Forest algorithm and a Support Vector Machine (SVM). For every image in the training dataset we extract the GLCM properties including \textit{contrast}, \textit{dissimilarity}, \textit{homogeneity}, \textit{ASM}, \textit{energy} and correlation for 4 different angles (rotationally invariant features) \textit{0, 45, 90, 135}. Additionally, to calculate those properties it is necessary to provide the distance given in pixels. In this experiment, we use distances of 1, 2 and 3 pixels. After calculating those features, we have a total of 72 features (6 properties x 4 angles x 3 distances = 72 features). In the statistical approach, it is clear that elements such as color or other important features of the images will be missing. However, if texture is important for CNN classifiers in general,
those same texture features might be sufficient to determine how well an image is recognizable by a classifier and if they are more resilient to attacks than other randomly selected images. 
\\
Table \ref{tab:models} shows the results of traditional machine learning (ML) approaches and deep learning models trained to determine whether an image is robust or non-robust. For the non-deep-learning approaches, 72 features are extracted using the statistical method GLCM. Despite the expected results of deep learning models performing better than traditional ML, the best performing ML model achieves 75 accuracy using only statistical features extracted from a gray-scale image. For the gray-scale image we used the Y channel from the YCbCr. According to Pestana et al. \cite{pestana2020adversarial}, the Y channel from YCbCr color space has more relevant features than the color CbCr channels. In addition, Geirhos et al.~\cite{geirhos_imagenet-trained_2019} discuss the importance of texture for CNN ImageNet-based classifiers. The results from Table \ref{tab:models} show that a model using extracted GLCM features from the Y-channel is able to recognise in the majority of cases whether an image is robust or not. An interesting point to note is that while deep models achieve higher accuracy using RGB images, the results in Table \ref{tab:models} also confirm that the most relevant information comes from the Y-channel and texture of an image. In Table \ref{tab:modelsunderattack}, we add PGD perturbations with $\epsilon = 0.01$ to the images and repeat the experiment of Table \ref{tab:models}. Our results show that the accuracy of deep models drops more drastically compared to the shallow models that use only statistical GLCM features from gray images for the same attack and amount of perturbation. This confirms that our simple feature based classifiers are good proxies to measure the robustness of a dataset.

% Table generated by Excel2LaTeX from sheet 'models-results'
\begin{table}[t]
  \centering
  \setlength\tabcolsep{1pt} % reduce blank spaces to half 3 pt (default 6 pts to fit table)
  \scriptsize
  \caption{Robust vs Non-robust classification of images under attack. The same classifiers from Table \ref{tab:models} are tested using adversarial images from a PGD attack $\epsilon$=0.01. Despite the initial accuracy of deep learning models, they seem to be much more susceptible than traditional ML classifiers using statistical GLCM features.}
    \begin{tabular}{|l|l|r|r|r|}
    \hline
     Model & Type Feature & \multicolumn{1}{l|}{Accuracy} & \multicolumn{1}{l|}{Precision} & \multicolumn{1}{l|}{~~Recall~} \\
    \hline
    Logistic Regression & Statistical GLCM & 63.70 & 72.2 & 52.80 \\
    Random Forest & Statistical GLCM & 73.08 & 75.80 & 77.30 \\
    SVM   & Statistical GLCM & 68.08 & 79.10 & 70.10 \\
    ResNet34 (Finetuned) & Deep Learning / CNN &    78.90 & 70.43 & 99.60 \\
    ResNet152 (Finetuned) & Deep Learning / CNN &    67.07 & 97.40 & 75.30 \\
    DenseNet121 (Finetuned) & Deep Learning / CNN &   67.29 & 97.00 & 75.70 \\
    \hline
    \end{tabular}%
  \label{tab:modelsunderattack}%
\end{table}%

\vspace{-1mm}
\section{Conclusion}
\vspace{-2mm}
 We showed the existence of \textit{defense-friendly} images that are both resilient (or robust) to adversarial attacks and also recover the under attack model's accuracy more easily than other images.  We demonstrated that the classification accuracy of models on images that are initially classified correctly by models of different architectures (but trained using the same ImageNet dataset) is easier to recover using a defense. This phenomenon may give the misleading impression regarding the performance of a given defense mechanism and has been exploited by some adversarial defense works which hand-pick a subset of clean images to report a near 100 defense accuracy. 
 However, Akhtar et al. \cite{akhtar2018defense} argue that evaluating defense mechanisms on already misclassified images is not meaningful and such images should not be considered in evaluation since an attack on a misclassified image is considered successful by default and this could mislead the interpretation of the results. We showed through extensive experimentation that such hand-picked datasets indeed give misleading advantages while evaluating the performance of a defense. To overcome this problem, we proposed three metrics to quantify the robustness/defense-friendliness of a dataset. We also provide a dataset with more than 15k robust images to complement the 7.5K natural adversarial examples of \cite{hendrycks2019natural}. We believe that our dataset and metrics will be valuable for unbiased benchmarking of defense mechanisms.
 \vspace{-1mm}
 
 \section{Acknowledgements}
  \vspace{-2.5mm}
 \noindent The main author was recipient of an Australian Government Research Training Program (RTP) Scholarship at The University of Western Australia. This research was supported by ARC Discovery Grant DP190102443. We would like to thank Robyn Owens for her insightful comments.

{\small
\bibliographystyle{ieee_fullname}
\bibliography{references}
}

\end{document}